\documentclass[10pt, a4paper]{article}

\usepackage{lrec2026} 

\usepackage{xcolor}
\usepackage{booktabs}
\usepackage{multirow} 
\usepackage{amsmath}

\title{Universal NER v2: Towards a Massively Multilingual Named Entity Recognition Benchmark}

\name{
\begin{tabular}{c}
Terra Blevins$^{1}$, Stephen Mayhew$^{2}$, Marek \v{S}uppa$^{3}$, Hila Gonen$^{4}$, Shachar Mirkin$^{5}$,\\
Vasile Pais$^{6}$, Kaja Dobrovoljc$^{7}$, Voula Giouli$^{8}$, Jun Kevin$^{9}$, \\
Eugene Jang$^{1}$, Eungseo Kim$^{10}$, Jeongyeon Seo$^{11}$, Xenophon Gialis$^{12}$, Yuval Pinter$^{13}$
\end{tabular}
}

\address{
\parbox{0.95\textwidth}{\centering
$^{1}$Northeastern University, USA \quad
$^{2}$Duolingo, USA \quad
$^{3}$Comenius University Bratislava, Slovakia \\
$^{4}$University of British Columbia, Canada \quad
$^{5}$Alpinference, France \quad
$^{6}$Research Institute for Artificial Intelligence, Romanian Academy, Romania \quad
$^{7}$University of Ljubljana, Slovenia \\
$^{8}$Aristotle University of Thessaloniki / ILSP, Athena Research Center, Greece \\
$^{9}$Universitas Pelita Harapan, Indonesia \quad
$^{10}$Seoul National University, South Korea \quad
$^{11}$Independent Researcher, South Korea \\
$^{12}$Democritus University of Thrace, Greece \quad
$^{13}$Ben-Gurion University of the Negev, Israel \\
\texttt{t.blevins@northeastern.edu, stephen@duolingo.com, marek@suppa.sk}, 
\texttt{hilagnn@gmail.com, shacharmirkin@gmail.com} \\
\texttt{ vasile@racai.ro, kaja.dobrovoljc@ff.uni-lj.si, pgiouli@del.auth.gr} \\
\texttt{junkevin88@gmail.com, ej16056@gmail.com} \\
\texttt{juniormoo@snu.ac.kr, yena.seo@kaist.ac.kr, xenogial@pme.duth.gr, yuvalpinter@gmail.com}
}
}

\abstract{
We present Universal NER (UNER) v2, a significant extension of the initial version released in 2024. UNER is a collaborative dataset for multilingual named-entity annotations, built to support research on NER methods in a cross-linguistic setting. UNER v2 adds 11 new datasets in 10 typologically varied language varieties to the resource, including multiple parallel evaluation benchmarks aligned with each other and other datasets in UNER v1, while maintaining the same annotation guidelines and high standards for inter-annotator agreement. We report detailed statistics for the dataset and benchmark UNER v2 using both encoder-based model architectures and LLMs. \\ \newline \Keywords{Named Entity Recognition, Multilingual, Benchmark, Cross-lingual Transfer, Large Language Models} }

\begin{document}

\maketitleabstract

\section{Introduction}
While multilingual language models promise to bring the benefits of LLMs to speakers of many languages, gold-standard evaluation benchmarks in most languages to interrogate these assumptions remain scarce. The Universal NER project, now entering its fourth year, is dedicated to building gold-standard multilingual Named Entity Recognition (NER) benchmark datasets. Inspired by existing massively multilingual efforts for other core NLP tasks \cite[e.g., Universal Dependencies;][]{de-marneffe-etal-2021-universal}, the project uses a general tagset and thorough annotation guidelines to collect standardized, cross-lingual annotations of named entity spans. The first installment (UNER v1) was released in 2024 \cite{mayhew-etal-2024-universal}, and the project has continued and expanded since then, with various organizers, annotators, and collaborators in an active community.

\begin{figure}[t!]
    \centering
    \includegraphics[width=\linewidth]{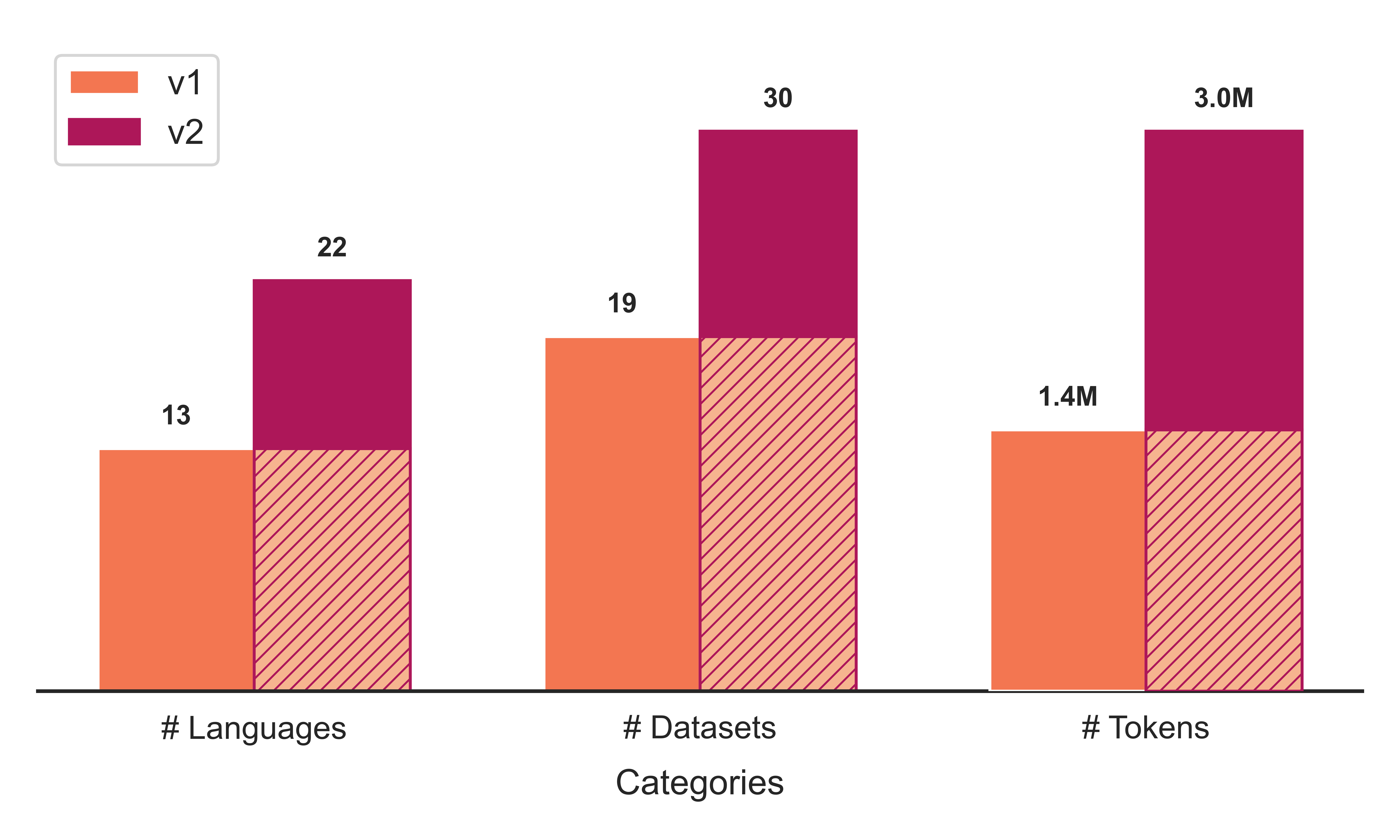}
    \caption{Comparison of the dataset statistics for UNER v1 and UNER v2.}
    \label{fig:overall-stats}
\end{figure}

\begin{table*}[t]
\centering
\begin{tabular}{ll ll}
\toprule
\multicolumn{2}{l}{Language} & \multicolumn{2}{l}{Dataset} \\
\midrule
\textsc{el} & Greek & \texttt{gdt} & Greek Dependency Treebank \\
\textsc{he} & Hebrew & \texttt{htb} & Hebrew Dependency Treebank \\
\textsc{nno} & Norwegian Nynorsk & \texttt{ndt} & Norwegian Dependency Treebank \\
\textsc{nob} & Norwegian Bokm{\aa}l & \texttt{ndt} & Norwegian Dependency Treebank \\
\textsc{sl} & Slovenian & \texttt{ssj} & Slovenian SSJ Treebank \\
\textsc{sv} & Swedish & \texttt{lines} & Swedish LinES Treebank \\
\midrule
\textsc{cs} & Czech & \texttt{pud} & Parallel Universal Dependencies \\
\textsc{id} & Indonesian & \texttt{pud} & Parallel Universal Dependencies \\
\textsc{ja} & Japanese & \texttt{pud} & Parallel Universal Dependencies \\
\textsc{ko} & Korean & \texttt{pud} & Parallel Universal Dependencies \\
\midrule
\textsc{ro} & Romanian & \texttt{legalnero} & Romanian Named Entities in the Legal Domain \\
\bottomrule
\end{tabular}
\caption{Languages and associated original dataset names used in UNER v2.}
\label{tab:data_fullnames}
\end{table*}

We present a substantial update to the Universal Named Entity Recognition (UNER) project, with new, gold-standard annotations on eleven datasets in ten languages (nine of which are new to the UNER collection), comprising 59,000 entities over 1.5 million tokens. As shown in \autoref{fig:overall-stats}, the addition of these datasets brings the composition of the overall Universal NER repository to 30 datasets across 22 languages, with a total of 3 million annotated tokens. We release this aggregated dataset as UNER v2.

In this work, we first provide a summary of the Universal NER project thus far (Section \ref{sec:bg}) and then describe the creation of UNER v2 datasets, along with relevant dataset statistics and analyses (Section \ref{sec:data}). We then present baseline experiments on the new datasets introduced in UNER v2, benchmarking them against both encoder-based and generative language models (Section \ref{sec:results}). Our experimental results show that state-of-the-art generative models achieve at most 0.50 average F1, with performance dropping sharply on typologically distant and lower-resourced languages. UNER continues to be an essential benchmark for developing more robust, truly multilingual NER systems.  

\section{Universal NER}
\label{sec:bg}

The development of multilingual Named Entity Recognition benchmarks has accelerated in recent years. The Universal NER project~\citep[UNER v1][]{mayhew-etal-2024-universal} established a community-driven gold standard resource in 13 languages, following the philosophy of projects such as Universal Dependencies~\citep[UD;][]{de-marneffe-etal-2021-universal} or the PARSEME initiative for multiword expressions \cite{savary2017parseme}.
Complementary efforts have also emerged in domain-specific contexts, such as LegalNERo \cite{pais2024legalnero}, which focuses on the Romanian legal domain, and NorNE \cite{jorgensen2020norne}, which contains named entity annotations for Norwegian.


The hallmark of the Universal NER project is a simple and transferable tagset consisting of 3 tags -- Person (\textsc{per}), Location (\textsc{loc}), and Organization (\textsc{org}) -- as well as thorough annotation guidelines. Annotation is performed using shared software based on TALEN~\cite{mayhew2018talen}, and gathered on GitHub. To encourage high-quality annotations, we require (with exceptions) that at least 2 annotators provide annotations on at least 5\% of the data, so that we can calculate agreement statistics. In many cases, we have found that even when there is low overlap between annotators, agreement statistics can uncover annotation guideline misunderstandings or incompetent annotators, and identify potential quality issues.

\paragraph{Impact of UNER v1}

Since its introduction in 2024, the first version of this dataset has had a substantial impact on the multilingual NLP community, garnering citations from over 40 research works. Its significance is further evidenced by its integration into major multilingual resources: the Aya Dataset, a prominent open-access collection for multilingual instruction tuning~\cite{aya}, and skLEP, a comprehensive Slovak language understanding benchmark~\cite{sklep}.

Other works have leveraged the first version of the dataset to investigate targeted questions—such as the effect of NER on translation~\cite{silp_nlp} and the efficacy of synthetic data for NER tasks~\cite{synthetic_ner}—as well as broader fundamental questions concerning cross-lingual transfer~\cite{liu2025alignment,cross_ie} and multilingual knowledge distillation~\cite{distillation}. The dataset continues to enable explorations across a wide spectrum of multilingual research directions \cite{impact1,impact2,impact3,impact4,impact5}.

\setlength{\tabcolsep}{4pt}
\begin{table*}[]
    \centering
    \scalebox{0.8}{
    \begin{tabular}{ll rrrr rrrr rrrr}
        \toprule
        &  & \multicolumn{4}{c}{Sentences} & \multicolumn{4}{c}{Entities} & \multicolumn{4}{c}{Tokens} \\
     \cmidrule(lr){3-6} \cmidrule(lr){7-10} \cmidrule(lr){11-14}
     Lang. & Dataset & Train & Dev & Test & All & Train & Dev & Test & All & Train & Dev & Test & All \\
\midrule
 \textsc{el} & \texttt{gdt} & 1,662 & 403 & 456 & 2,521 & 1,551 & 501 & 436 & 2,488 & 42,326 & 10,443 & 10,672 & 63,441 \\ 
 \textsc{he} & \texttt{htb} & 5,241 & 484 & 491 & 6,216 & 6,013 & 434 & 439 & 6,886 & 137,717 & 11,412 & 12,282 & 161,411 \\
 \textsc{nno} & \texttt{ndt} & 14,174 & 1,890 & 1,511 & 17,575 & 10,348 & 1,111 & 897 & 12,356 & 245,330 & 31,250 & 24,773 & 301,353 \\
 \textsc{nob} & \texttt{ndt} & 15,696 & 2,409 & 1,939 & 20,044 & 10,062 & 1,438 & 1,259 & 12,759 & 243,886 & 36,369 & 29,966 & 310,221 \\
 \textsc{sl} & \texttt{ssj} & 10,903 & 1,250 & 1,282 & 13,435 & 7,293 & 907 & 724 & 8,924 & 215,155 & 26,500 & 25,442 & 267,097 \\
 \textsc{sv} & \texttt{lines} & 3,176 & 1,032 & 1,035 & 5,243 & 1,221 & 433 & 451 & 2,105 & 55,451 & 18,515 & 16,994 & 90,960 \\
 \midrule
 \textsc{cs} & \texttt{pud} & -- & -- & 1,000 & 1,000 & -- & -- & 1,008 & 1,008 & -- & -- & 18,610 & 18,610 \\
 \textsc{id} & \texttt{pud} & -- & -- & 1,000 & 1,000 & -- & -- & 1,110 & 1,110 & -- & -- & 19,446 & 19,446 \\
 \textsc{ja} & \texttt{pud} & -- & -- & 1,000 & 1,000 & -- & -- & 1,150 & 1,150 & -- & -- & 28,788 & 28,788 \\
 \textsc{ko} & \texttt{pud} & -- & -- & 1,000 & 1,000 & -- & -- & 1,172 & 1,172 & -- & -- & 16,584 & 16,584 \\
 \midrule
 \textsc{ro} & \texttt{legalnero} & -- & -- & 8,284 & 8,284 & -- & -- & 8,996 & 8,996 & -- & -- & 265,335 & 265,335 \\
\bottomrule
    \end{tabular}
    }
    \caption{Dataset statistics for the new benchmarks included in UNER v2.}
    \label{tab:data}
\end{table*}

\section{UNER v2 Creation and Dataset}
\label{sec:data}
Universal NER (UNER) v2 is an extension of UNER that adds 11 new datasets in 10 new language varieties (with Norwegian represented by two written standards, Nynorsk and Bokmål), namely Modern Greek, Hebrew, Norwegian Nynorsk, Norwegian Bokmål, Slovenian, Swedish, Czech, Indonesian, Japanese, Korean, and Romanian  (\autoref{tab:data}; the full data composition of UNER v2 is given in \autoref{fig:overall-stats}), for a total of 30 datasets spanning 22 languages now included in the UNER project.
This updated version of UNER also incorporates minor annotation fixes for some existing datasets originally released in v1 (namely, English's \texttt{EWT} and \texttt{PUD}).
Here, we present the annotation process for UNER v2 (Section \ref{sec:data-annot}) and the dataset statistics for the new datasets released in v2 (Section \ref{sec:data-stats}). Additionally, we conduct a cross-lingual analysis of the \texttt{pud} labels to evaluate inter-language consistency in named entity usage across parallel datasets (Section \ref{sec:data-pud}).

\subsection{Data Annotation}
\label{sec:data-annot}
The annotation process for UNER v2 generally followed the procedure laid out in the first UNER data collection effort. We used the same annotation guidelines, with minor wording changes to clarify the class distinctions: 
documents are annotated for location (\textsc{loc}), organization (\textsc{org}), and person (\textsc{per}) entities.\footnote{While entities that do not fall into these categories are labeled as \textsc{oth} (Other) during annotation, these additional entities are not included in the final dataset release.} 
Annotators for each dataset were recruited as volunteers from the NLP community, and the annotation effort was primarily coordinated through a collaborative Discord server.

Following UNER v1, annotations were collected using the TALEN platform \cite{mayhew2018talen}, a web-based tool developed for span-level sequence labeling. Additionally, all datasets annotated from scratch for UNER v2 include overlapping annotations from multiple annotators for at least 5\% of dataset documents to calculate inter-annotator agreement (\autoref{tab:iaa}). Each dataset is released in a custom \texttt{.conllu} format (termed \texttt{.iob2}); the overall format follows \texttt{.conllu}~\cite{nivre-etal-2020-universal} but contains the following column types for word-level information: word id, word form, UNER label, XNER label,\footnote{XNER is only present for datasets that have been transferred to the UNER format, and contains the NER labels from the original datasets.} and annotator id. Since text is annotated at the word-level in this format, NER spans are annotated using the same IOB2 annotation schema as the datasets in v1. 

The newly added Greek (\textsc{el}) dataset was annotated on top of the Greek \texttt{gdt} \cite{prokopidis2017universal} in accordance with the UNER guidelines. Language-specific examples from prior -- yet compatible -- annotation efforts in the domains of finance \cite{boutsis2000system} and in multi-domain journalistic texts \cite{giouli-etal-2006-multi} were actively consulted during annotation to guide challenging cases and support consistent decisions.

\paragraph{Dataset Transfer} UNER v2 also incorporates three existing datasets converted to align with the UNER annotation standard. Norwegian \texttt{ndt} (containing two written standards \textsc{nno} and \textsc{nob}) and Slovenian (\textsc{sl}) \texttt{ssj} are originally annotated in existing Universal Dependencies treebanks~\cite{ovrelid-hohle-2016-universal,dobrovoljc-etal-2017-universal}. Norwegian \texttt{ndt} is annotated with named entities as an independent effort by~\citet{jorgensen2020norne}, and then automatically converted to the UNER format with a fixed label mapping. The Slovenian \texttt{ssj} dataset, by contrast, contains partial named entity annotations based on the JANES-NER scheme~\cite{arhar-holdt-etal-2024-suk,zupan2017_janesner_guidelines}. The conversion to UNER completed these annotations by adding 1,383 new entities and then automatically converting and manually correcting all labels to conform to the UNER guidelines.

However, Romanian (\textsc{ro}) \texttt{legalnero} is not part of UD and is instead annotated on a dataset of Romanian legal documents~\cite{pais2024legalnero}. The original version provides gold annotations for organizations, locations, persons, time expressions, and legal resources mentioned in legal documents. The raw text files were extracted from the Romanian part~\citep{tufi-EtAl:2020:LREC} of the MARCELL corpus~\citep{varadi-EtAl:2020:LREC}. 
The annotations are automatically converted to the UNER format by selecting only the entity types supported by the UNER scheme. The corpus contains 265,335 tokens in 8,284 sentences over 370 text files.

\begin{table*}
    \centering
    \scalebox{0.8}{
    \begin{tabular}{l ll l rrr}
    \toprule
    & & & & \multicolumn{3}{c}{Entity Dist. (\%)} \\
    \cmidrule(lr){5-7}
    Data Source & Lang. & Dataset & Domains & \texttt{LOC} & \texttt{ORG} & \texttt{PER}\\
    \midrule
    \citet{prokopidis2017universal} & \textsc{el} & \texttt{gdt} & news, spoken, wiki & 38.0\% & 41.5\% & 20.5\% \\
    \citet{tsarfaty2013unified, mcdonald2013universal} & \textsc{he} & \texttt{htb} & news & 24.0\% & 39.7\% & 36.3\% \\
    \multirow{2}{*}{\citet{ovrelid-hohle-2016-universal}} & \textsc{nno} & \texttt{ndt} & blog, news, nonfiction & 28.4\% & 30.1\% & 41.5\% \\
     & \textsc{nob} & \texttt{ndt} & blog, news, nonfiction & 27.3\% & 31.6\% & 41.1\% \\
    \citet{dobrovoljc-etal-2017-universal} & \textsc{sl} & \texttt{ssj} & fiction, news, nonfiction & 34.1\% & 22.1\% & 43.8\% \\
    \citet{ahrenberg2015converting} & \textsc{sv} & \texttt{lines} & fiction, nonfiction, spoken & 18.5\% & 7.9\% & 73.6\% \\
     \midrule
     \multirow{4}{*}{\citet{zeman2018conll}}& \textsc{cs} & \texttt{pud} & news, wiki & 38.3\% & 20.8\% & 40.9\% \\
     & \textsc{id} & \texttt{pud} & news, wiki & 47.0\% & 16.1\% & 36.9\% \\
     & \textsc{ja} & \texttt{pud} & news, wiki & 47.0\% & 16.9\% & 36.1\%  \\
     & \textsc{ko} & \texttt{pud} & news, wiki & 40.0\% & 24.7\% & 35.3\% \\
     \midrule    
     \citet{pais2024legalnero} & \textsc{ro} & \texttt{legalnero} & legal & 21.7\% & 68.4\% & 9.9\% \\
    \bottomrule
    \end{tabular}}
    \caption{Domains and distribution of entity types for datasets in UNER v2. Domains are categorized for the underlying UD datasets at \url{https://universaldependencies.org/} or author descriptions if not included in UD.}
    \label{tab:uner-domains}
\end{table*}

\paragraph{Annotation Corrections} In addition to adding new datasets to Universal NER, UNER v2 also comprises revisions of the English \texttt{EWT} and \texttt{PUD} datasets. Specifically, we relabel a handful of entities ($\leq3$) in these datasets that were found to be incorrect in the v1 version. The total number of annotated entities remains the same. 

\subsection{Dataset Statistics}
\label{sec:data-stats}
\autoref{tab:data} illustrates the dataset statistics for each of the new datasets released in UNER v2. This new version of UNER adds six full NER datasets, as well as four new parallel NER evaluation sets that are aligned with six additional languages already annotated in UNER v1. The new datasets are also typologically diverse, spanning five macro-language families (Indo-European, Afroasiatic, Austronesian, Koreanic, and Japonic); this brings the full coverage of the UNER datasets to 22 languages across six language families.\footnote{UNER v1 includes the Sino-Tibetan family with two Chinese datasets.}
The new languages include fusional (Czech, Greek, Norwegian,
Slovenian, Swedish), agglutinative (Japanese, Korean), analytic
(Indonesian), and introflexive (Hebrew) morphological types, written
in four scripts (Latin, Greek, Hebrew, and CJK). This diversity
matters for NER: morphological complexity affects where entity
boundaries fall (e.g., agglutinative suffixes in Korean), and script
differences impact how multilingual models tokenize text.

In addition to linguistic diversity, the datasets in UNER v2 are also stylistically diverse (\autoref{tab:uner-domains}). The datasets broadly cover fiction and nonfiction text, as well as spoken-language transcripts. Additionally, \texttt{legalnero} adds the legal domain to UNER in Romanian, while \textsc{sv} \texttt{lines} expands UNER's Swedish coverage to include literary texts. We correspondingly observe distribution shifts in entity use across domains; for example, the legal documents in \texttt{legalnero} contain a much higher proportion of \textsc{org} entities than the other datasets, while having very few ($<10\%$) \textsc{per} entities.

Finally, we present the inter-annotator agreement (IAA) for new datasets in UNER v2 in \autoref{tab:iaa}. 
We observe similar agreement trends to those seen in the original UNER annotation process: specifically, agreement between the annotators on \textsc{per} entities is usually much higher than on the \textsc{loc} and \textsc{org} entities, due to the prevalence of ambiguous entities that depend on (often vague or missing) context. For instance, metonymic organizations are common, leading to the use of the same form to describe an organization (\textit{The White House} announced a new policy today) and a location (The meeting was held at \textit{the White House}). 

\subsection{Analyzing Named Entities across Parallel Multilingual Data}
\label{sec:data-pud}
A key contribution of Universal NER is the creation of parallel gold-standard evaluation benchmarks for Named Entity Recognition, annotated on top of Parallel UD~\cite[PUD;][]{zeman2018conll}; this release of UNER adds four more typologically diverse languages to this resource: Czech (fusional SVO), Indonesian (analytic SVO), and Japanese and Korean (agglutinative SOV). In this section, we examine the cross-lingual consistency of named entity usage across all ten languages in the UNER dataset. 

\begin{figure*}
    \centering
    \includegraphics[width=\linewidth]{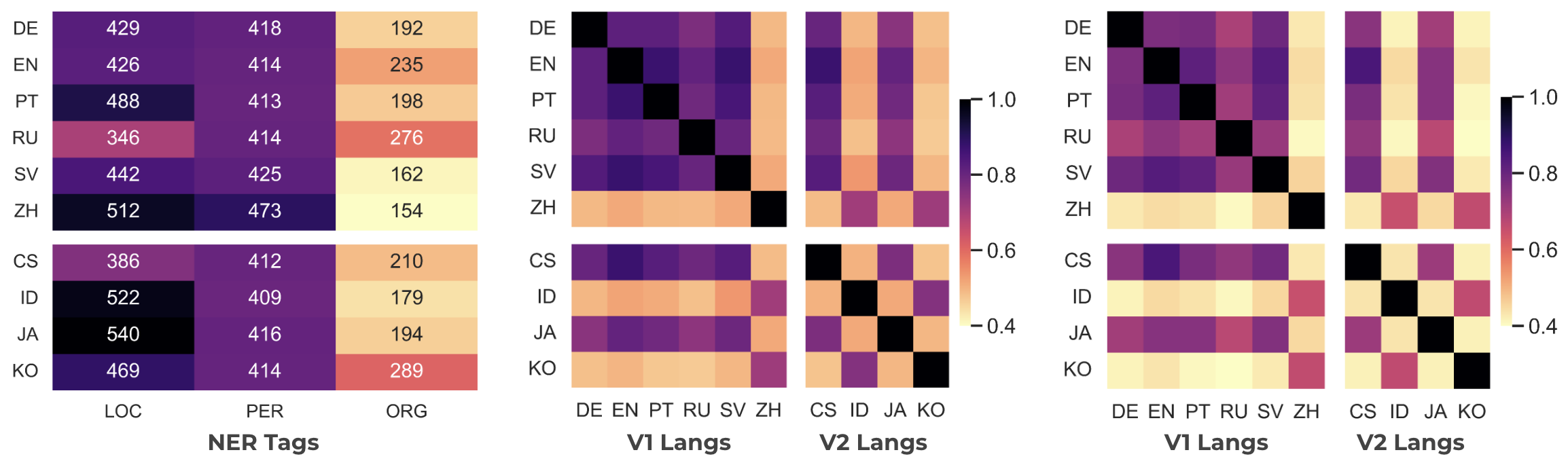}
    \caption{A cross-lingual comparison of UNER annotations on top of parallel text (PUD). We consider the overlap of datasets from UNER v1 and v2. \textbf{Left}: The overall tag distribution in each PUD dataset. \textbf{Center}: Sentence-level agreement between languages for entity count. \textbf{Right}: Sentence-level agreement on entity count within each class (\textsc{loc}, \textsc{org}, \textsc{per}) between languages.}
    \label{fig:pud-comp}
\end{figure*}

\autoref{fig:pud-comp} displays the results of our cross-dataset analysis. Similar to the initial analysis of the UNER v1 parallel datasets, the Indo-European (IE) languages (including the addition of Czech), strongly agree in terms of entity distribution and use. In contrast, by adding examples from other families, we observe that the behavior of languages outside the IE family is more complex. Indonesian and Korean behave similarly to Chinese, with less overlap with IE languages (and moderate agreement with each other and with Chinese); in contrast, Japanese exhibits alignment patterns similar to those of the IE languages, albeit with many additional \textsc{loc} entities.

\begin{table*}[]
    \centering
    \scalebox{0.8}{
    \begin{tabular}{ll rrrr rrrr rrrr}
\toprule
& & \multicolumn{4}{c}{Train} & \multicolumn{4}{c}{Dev} & \multicolumn{4}{c}{Test} \\
\cmidrule(lr){3-6} \cmidrule(lr){7-10} \cmidrule(lr){11-14}
Lang. & Dataset & \texttt{LOC} & \texttt{ORG} & \texttt{PER} & \% Docs & \texttt{LOC} & \texttt{ORG} & \texttt{PER} & \% Docs & \texttt{LOC} & \texttt{ORG} & \texttt{PER} & \% Docs \\
\midrule
 \textsc{el} & \texttt{gdt} & 0.750 & 0.556 & 0.962 & 25\% & 0.843 & 0.607 & 0.883 & 25\% & 0.750 & 0.493 & 0.923 & 26.9\% \\ 
 \textsc{he} & \texttt{htb} & 0.777 & 0.785 & 0.839 & 29.7\% & 0.741 & 0.664 & 0.968 & 100\% & 0.746 & 0.831 & 0.914 & 100\% \\
 \textsc{sv} & \texttt{lines} & ** & ** & ** & ** & 0.927 & 0.526 & 0.974 & 100\% & 0.571 & 0.667 & 0.962 & 25\% \\
 \midrule
 \textsc{cs} & \texttt{pud} & -- & -- & -- & -- & -- & -- & -- & -- & 0.656 & 0.840 & 0.775 & 5.3\% \\
 \textsc{id} & \texttt{pud} & -- & -- & -- & -- & -- & -- & -- & -- & 0.629 & 0.590 & 0.930 & 8.8\% \\ 
 \textsc{ja} & \texttt{pud} & -- & -- & -- & -- & -- & -- & -- & -- & 0.651 & 0.769 & 0.839 & 6.3\% \\
\textsc{ko} & \texttt{pud} & -- & -- & -- & -- & -- & -- & -- & -- & 0.613 & 0.443 & 0.945 & 11.8\% \\
\bottomrule
    \end{tabular}
    }
    \caption{Inter-annotator agreement (IAA) for the new benchmarks included in UNER v2. -- indicates that the data split does not exist, while ** indicates that IAA was not collected. IAA was also not calculated for datasets transferred into UNER v2: \textsc{nno} and \textsc{nob} \texttt{ndt}, \textsc{sl} \texttt{ssj}, and \textsc{ro} \texttt{legalnero}.}
    \label{tab:iaa}
\end{table*}

\section{Experiments}
\label{sec:results}
We perform two sets of experiments: (a) using a finetuned cross-lingual encoder, XLM-R~\cite{conneau2020unsupervised}, to be comparable with our prior work, Section \ref{sec:baselines}; (b) by directly prompting three large language models (LLMs), as detailed in Section \ref{sec:llm_res}.

\subsection{Traditional baselines}

\label{sec:baselines}

\paragraph{Experiment Setup}
This section establishes baselines on the new datasets in UNER v2 and provides in-language and cross-lingual results with XLM-R$_{\text{Large}}$. We finetuned XLM-R$_{\text{Large}}$ (560M parameters)~\cite{conneau2020unsupervised} on the UNER datasets in which train and dev sets are available,\footnote{Greek (\texttt{el\_gdt}), Hebrew (\texttt{he\_htb}), Norwegian Nynorsk (\texttt{nno\_ndt}), Norwegian Bokmål (\texttt{nob\_ndt}), Slovenian (\texttt{sl\_ssj}), Swedish (\texttt{sv\_lines}), English (\texttt{en\_ewt})} using a single NVIDIA GeForce RTX 3090 GPU. We used a learning rate of 3e-5 and batch size of 8, except for Korean (\texttt{kor\_pud}), where we used a batch size of 4. All code was adapted from the Huggingface transformers package~\cite{wolf2020transformers}.

\begin{figure*}[]
    \centering
    \includegraphics[width=0.9\linewidth]{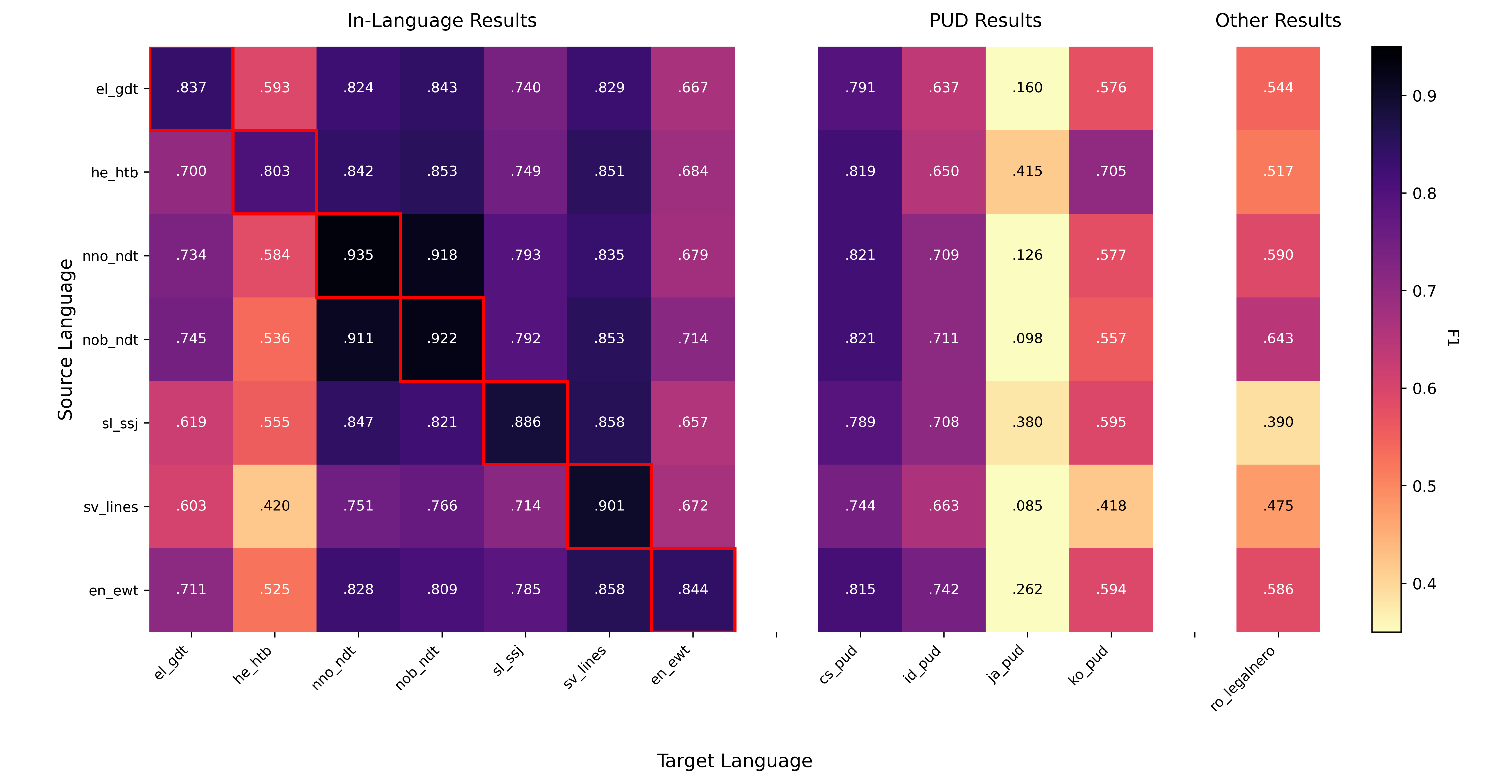}
    \caption{Experimental results for in-language and cross-lingual UNER performance with XLM-R$_{\text{large}}$.}
    \label{fig:experiments}
\end{figure*}

\paragraph{Results and Discussion}
Figure \ref{fig:experiments} reports the micro F1 scores on all test sets when XLM-R$_{\text{Large}}$ is finetuned on different languages. The in-language performance shown on the diagonal is almost always the highest among all test sets, with a few exceptions for closely related languages. We observe that in most cases, cross-lingual transfer performs well between European languages, achieving over 0.60 F1. However, transfer results in lower performance on non-European languages such as Japanese (\texttt{ja\_pud}) and Korean (\texttt{ko\_pud}), which aligns with observations from previous work that cross-lingual transfer to typologically distant languages remains challenging~\cite{chen2023a,wu2020a}.

Overall, the tag-level performance breakdown reveals that F1 on \textsc{org} is consistently the lowest, and \textsc{loc} is often the second lowest. This is consistent with the ambiguity between \textsc{org} and \textsc{loc} entities (e.g., metonymic uses discussed in Section \ref{sec:data-stats}), whereas person names are usually less ambiguous, resulting in a higher F1 on \textsc{per} for most datasets.

\subsection{LLM Results}
\label{sec:llm_res}

LLMs have been shown to be beneficial for text annotation tasks, particularly for Named Entity Recognition~\cite{tan2024llm, wang2023gptner}. Recent studies have explored their potential as cost-effective annotators and data generators~\cite{dagdelen2024structured, bogdanov2024nuner}, though their performance typically falls short of fine-tuned models~\cite{hu2024improving}.

We ran a set of experiments with LLMs, providing the model with the annotation guidelines and asking it to follow them when annotating each sentence in the dataset. 
We experimented with three models as our LLM annotator: \textit{Gemini 2.5 Flash Lite} (\textit{Gemini} below), \textit{Claude Sonnet 4} (\textit{Claude}), and \textit{GPT-5 Mini} (\textit{GPT} below).\footnote{Exact model versions: \texttt{gemini-2.5-flash-lite}, \texttt{anthropic.claude-sonnet-4-20250514-v1}, \texttt{gpt-5-mini-2025-08-07}.} These were chosen to represent the balance between state-of-the-art performance and inference cost at the time of writing. We deliberately did not engage in prompt engineering, as we aimed to observe what LLMs can achieve when presented with the exact same guidelines as human annotators received. In that, we can consider the results shown here as a baseline for LLM-based annotations on this dataset.

\begin{figure}
    \centering
    \includegraphics[width=\linewidth]{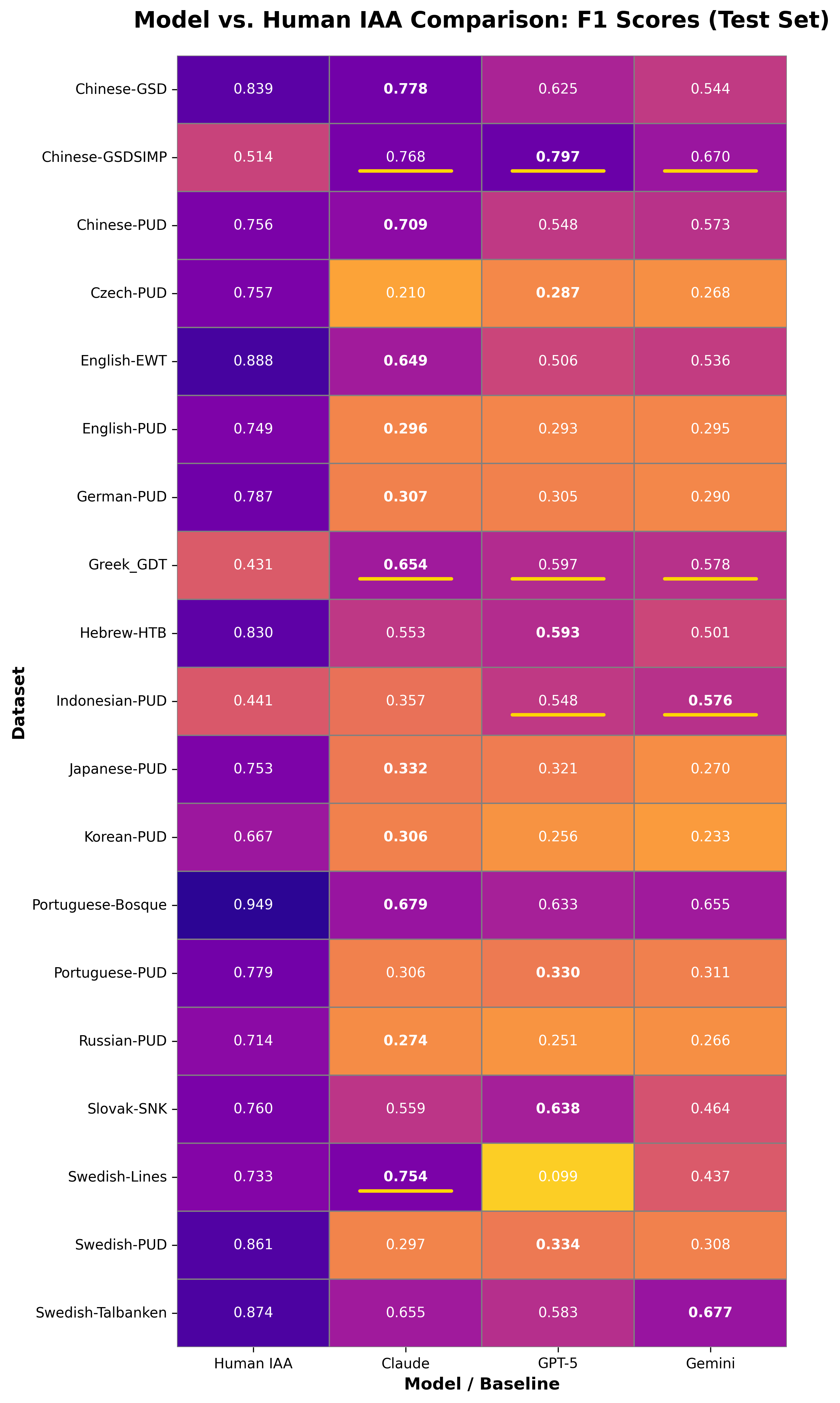}
    \caption{F1 score comparison across 19 multilingual NER datasets (test sets) for three LLMs against human Inter-Annotator Agreement (IAA) baseline. Bold values indicate the best-performing model for each dataset; underlined values indicate scores exceeding human agreement.}
    \label{fig:llm-results}
\end{figure}

Figure~\ref{fig:llm-results} presents the inter-annotator agreement between each model and the human annotator who annotated the most documents in a given dataset. For brevity, we include a sample of the results, specifically removing datasets that are too small to be considered meaningful for this measurement. The left-most column of the table shows the human IAA for comparison.

Among the three models, Claude achieves the highest average F1 (0.497), followed by GPT and Gemini, though all remain substantially below human inter-annotator agreement (0.741 average F1).
Performance varies across languages, from strong agreement on Greek \texttt{el\_gdt} to low agreement on Czech \texttt{cs\_pud}.\footnote{We note that the agreement depends not only on the language and the dataset but also on the human annotators themselves.} As with human IAA scores, \textsc{per} is the easiest entity type for LLMs to annotate, while \textsc{loc} and \textsc{org} are harder.

A closer look at annotation counts reveals systematic over- and under-annotation patterns. On the PUD parallel datasets, GPT and Gemini consistently produce 30--90\% more entity annotations than humans (e.g., on English PUD, humans annotate 682 entities while GPT produces 1,262), suggesting widespread annotation of non-\textbf{named} entities such as generic references to organizations. This pattern is language-specific: on Hebrew and Korean, all three models instead \textit{under}-annotate, producing 30--60\% fewer entities than humans. We also observed that models overlooked nuances in the guidelines, such as the requirement to annotate geopolitical entities as organizations rather than locations, and the distinction between nationalities and locations.


Although LLM annotation quality falls behind that of humans on this dataset, it can likely be improved with prompt engineering or an agentic flow (equivalent to a human annotator discussing a specific annotation). We note that LLMs have potential not only as annotators, but also as a means to surface errors in human annotations and to identify flaws in the annotation guidelines. We intend to explore these directions in the future.

\section{Prior Work}

In parallel to the UNER efforts, new multilingual datasets are being created to support broader instruction tuning and evaluation. The Aya dataset~\cite{aya} provides large-scale multilingual resources for instruction tuning, while \citet{zhang2024ner} propose a classification framework to better organize and understand the diversity of NER datasets. Beyond traditional NER, benchmarking efforts have expanded to more complex evaluation settings, such as situational awareness of large language models~\cite{tang2024situational}.

On the modeling side, advances in multilingual representation learning have also influenced NER. Recent work has investigated middle-layer alignment for cross-lingual transfer in fine-tuned large language models~\cite{liu2025alignment}, meta-pretraining strategies for zero-shot NER in low-resource Philippine languages~\cite{impact1}, and dynamic tokenization approaches for retrofitting large language models~\cite{impact2}. Novel paradigms such as multilingual pretraining for pixel language models~\cite{impact3} also highlight emerging directions beyond conventional text-only approaches.

\section{Conclusion}

We presented Universal NER v2, a new and substantially expanded version of the ongoing Universal NER project.
We are excited to see the steady pace at which the resource is growing, and hope that it can reach the magnitude of other massive endeavors in multilingual linguistic annotation such as Universal Dependencies and UniMorph.
As we release v2, significant work is being done towards incorporating more languages in the resource, including all levels from annotation to validation.
Through its coverage of less-resourced or even underrepresented and typologically diverse languages, the dataset contributes to a more inclusive and universal approach to language technology.
We invite more collaborators to contribute to future versions, whether in existing or new languages. 

UNER v1 has already made an impact on the NER community, facilitating multilingual and cross-lingual evaluation of modeling techniques in a controlled, and even parallel setting.
UNER v2 enhances and furthers this core ability. We eagerly anticipate the further advances in multilingual understanding that this will facilitate.


\section*{Data and Code Availability}
The UNER v2 dataset, annotation guidelines, and benchmarking code are publicly available.
The project website,\footnote{\url{https://www.universalner.org/}} including annotation
guidelines,\footnote{\url{https://www.universalner.org/guidelines/}} provides an overview
of the resource. All annotations are released under the CC-BY-SA-4.0 license and hosted
on GitHub.\footnote{\url{https://github.com/UniversalNER}} The dataset is also available
on Hugging Face.\footnote{\url{https://huggingface.co/datasets/universalner/universal_ner}}

\section*{Acknowledgments}
Multiple datasets new to UNER 2.0 were developed thanks to collaboration via the CA21167 COST action UniDive, funded by COST (European Cooperation in Science and Technology). Slovenian dataset creation was partially supported by grants ARIS-GC-002 and HORIZON-WIDERA-2023-TALENTS-01-01-101186647. M\v{S} was partially funded by the EU NextGenerationEU through the Recovery and Resilience Plan for Slovakia under the project No. 09I02-03-V01-00029.

\section{References}\label{sec:reference}

\bibliographystyle{lrec2026-natbib}
\bibliography{lrec2026-example}


\appendix

\end{document}